\documentclass{article}
\usepackage{ijcai22}

\usepackage{times}
\usepackage{soul}
\usepackage{url}
\usepackage[hidelinks]{hyperref}
\usepackage[utf8]{inputenc}
\usepackage[small]{caption}
\usepackage{graphicx}
\usepackage{amsmath}
\usepackage{amsthm}
\usepackage{bm}
\usepackage{multirow}
\usepackage{booktabs}
\usepackage{algorithm}
\usepackage{algorithmic}
\usepackage{color}
\title{Representation Learning for Compressed Video Action Recognition via Attentive Cross-modal Interaction with Motion Enhancement}

\author{
Bing Li$^{1,2}$
\and
Jiaxin Chen$^2$\footnotemark[1]\and
Dongming Zhang$^{3}$\and
Xiuguo Bao$^{3}$\and
Di Huang$^{1,2}$
\affiliations
$^1$State Key Laboratory of Software Development Environment, Beihang University, Beijing, China\\
$^2$School of Computer Science and Engineering, Beihang University, Beijing, China\\
$^3$National Computer Network Emergency Response Technical Team, Coordination Center of China, Beijing, China\\
\emails
\{libingsy, jiaxinchen, dhuang\}@buaa.edu.cn,~zhdm@cert.org.cn,~ baoxiuguo@139.com
}

\begin{document}

\maketitle

\begin{abstract}
    Compressed video action recognition has recently drawn growing attention, since it remarkably reduces the storage and computational cost via replacing raw videos by sparsely sampled RGB frames and compressed motion cues (\emph{e.g.}, motion vectors and residuals). However, this task severely suffers from the coarse and noisy dynamics and the insufficient fusion of the heterogeneous RGB and motion modalities. To address the two issues above, this paper proposes a novel framework, namely Attentive Cross-modal Interaction Network with Motion Enhancement (MEACI-Net). It follows the two-stream architecture, \emph{i.e.} one for the RGB modality and the other for the motion modality. Particularly, the motion stream employs a multi-scale block embedded with a denoising module to enhance representation learning. The interaction between the two streams is then strengthened by introducing the Selective Motion Complement (SMC) and Cross-Modality Augment (CMA) modules, where SMC complements the RGB modality with spatio-temporally attentive local motion features and CMA further combines the two modalities with selective feature augmentation. Extensive experiments on the UCF-101, HMDB-51 and Kinetics-400 benchmarks demonstrate the effectiveness and efficiency of MEACI-Net.
    \end{abstract}
    
    \section{Introduction}
    \renewcommand{\thefootnote}{\fnsymbol{footnote}}
    \footnotetext[1]{Corresponding author.}
    Along with the development of web applications and innovation of imaging sensors, the amount of videos is explosively growing, and automatic analysis of video content has become extremely important. Action recognition is a fundamental topic and has recently received increasing attention within the community. The past two decades have witnessed the advances successively by hand-crafted features and deep Convolutional Neural Networks (CNNs). The majority of the efforts are made to capture both static and dynamic cues from raw videos with all RGB frames decoded \cite{wang2020tdn}, and report excellent performance on the public benchmarks (\emph{e.g.} UCF-101 and HMDB-51). However, they generally require optical flows to represent motions, thus suffering from the drawbacks of high storage consumption and slow processing speed, and hardly meet the real-world demands.
    
    \begin{figure}[t]
        \centerline{\includegraphics[width=\linewidth, height=1.4in]{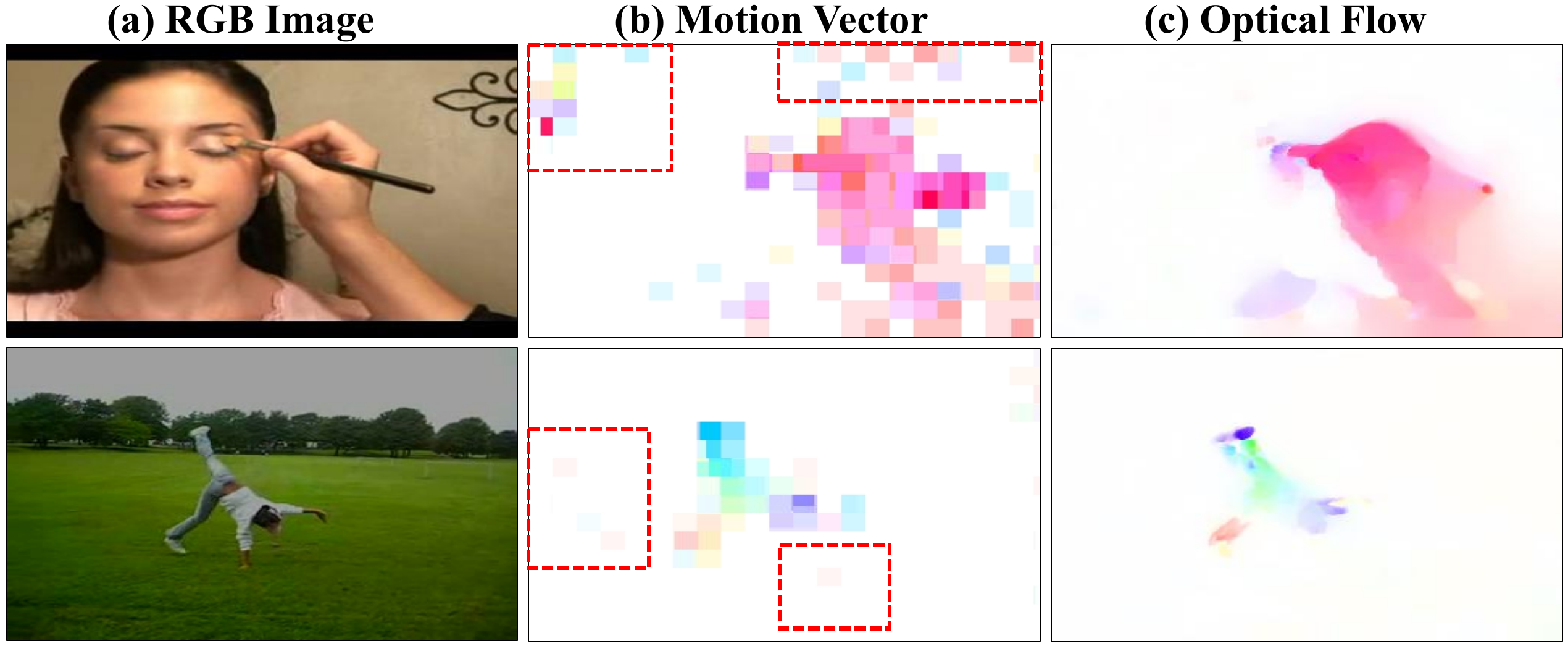}}
        \caption{(b) and (c) show coarse and noisy motion vectors compared to optical flows. Images in one row correspond to the same frame. Interferential backgrounds are highlighted by red rectangles.}
        \label{fig:1}
    \end{figure}

    More recently, action recognition in compressed videos is investigated as a promising alternative to handle the challenges above. In compressed videos, only a small proportion of frames are fully decoded, known as intra-frames (I-frames), and most frames are incompletely decoded, known as predicted frames (P-frames), which convey free but coarse motion information. I-frames, motion vectors and residuals are usually employed in compressed video based methods to replace decoded RGB frames and computed optical flows, where two major issues are critical, \emph{i.e.} motion enhancement and multi-modal fusion.
    \footnotetext[2]{The paper with \emph{Supplementary Material} is available at https://arxiv.org/abs/2205.03569.}
    
    On the one hand, those motion patterns recorded in compressed videos are not so accurate as those in raw videos as shown in Fig.~\ref{fig:1}, because the spatial resolution of the motion vector is substantially reduced (\emph{i.e.} 16$\times$) by block-wise matching and noise is inevitably induced by operations such as signal quantization and motion estimation in the encoding phase. For motion enhancement, taking optical flows as templates, \cite{shou2019dmc} applies Generative Adversarial Networks (GAN) to refine motion vectors, and \cite{cao2019compressed} removes unreliable movements in motion vectors according to a fixed threshold. Such techniques indeed contribute, but they update motion vectors or residuals in the input side and process them by the models designed for raw videos, which are not fully optimized in an end-to-end way, limiting further performance gains.
    
    On the other hand, to build comprehensive representation, it is required to integrate the information provided by the multiple modalities in compressed videos, \emph{i.e.} I-frames and P-frames (motion vectors and residuals). Current studies do not pay adequate attention to this step and generally apply simple early- or late-fusion as on raw videos. Unfortunately, I-frames and P-frames are produced at varying sampling rates, and such an inconsistency tends to impair the complementary among modalities. \cite{huo2019mobile} designs Temporal Trilinear Pooling to induce temporal contexts, which makes use of feature ensemble on corresponding frames of different modalities as well as another following I-frame for prediction. Although this method proves effective, it lacks local cross-modal interaction especially in the low-level layers, leaving much room for improvement.
    
    In this paper, we propose a novel approach to action recognition in compressed videos, namely Attentive Cross-modal Interaction Network with Motion Enhancement (MEACI-Net), which simultaneously addresses the problems of motion enhancement and multi-modal fusion. MEACI-Net follows the classic two-stream framework, for RGB (I-frames) and motion (P-frames) respectively. Considering the coarse and noisy nature of dynamics, we introduce a multi-scale building block in the Compressed Motion Enhancement (CME) network with a denoising module embedded, which guides the model to attend task-relevant spatial regions, resulting in more discriminative features to describe actions. Besides, we present the Selective Motion Complement (SMC) and Cross-Modality Augment (CMA) modules to learn local and global cross-modal interactions, thus facilitating feature fusion between modalities. We extensively evaluate MEACI-Net on three benchmarks, \emph{i.e.} HMDB-51, UCF-101 and Kinetics-400, and achieve state-of-the-art accuracies of action recognition on compressed videos, significantly reducing the gap to the ones on raw videos.
    
    \begin{figure*}[t]
        \centerline{\includegraphics[width=\textwidth]{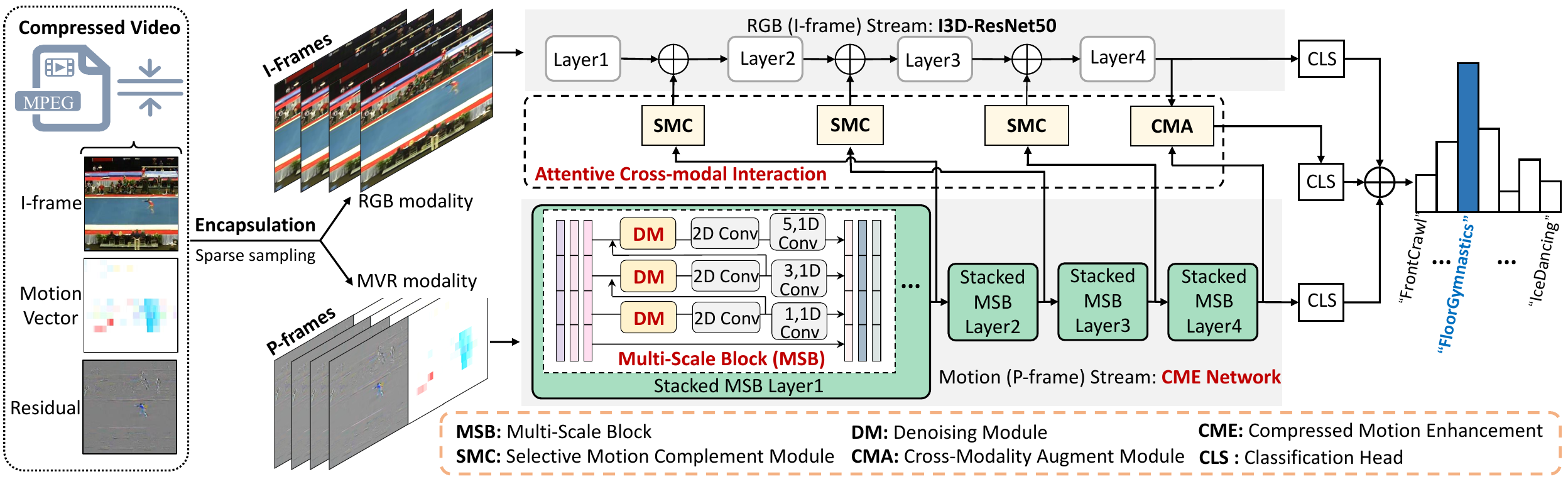}}
        \caption{Framework of the proposed MEACI-Net. We encapsulate I-frame, Motion Vector and Residual (MVR) from compressed videos and constitutes the RGB and MVR modalities, respectively. I3D-ResNet50 is directly employed to process I-frame clips while the CME network is designed to work with P-frame clips. The SMC and CMA facilitate cross-modal interaction in feature fusion from two modalities.}
        \label{fig:2}
    \end{figure*} 
    
    \section{Related Work}
    
    \subsection{Action Recognition in Raw Videos}
    
    Action recognition has been extensively investigated during the past two decades. Early works employ hand-crafted features \cite{wang2013action} and the recent ones build deep representations \cite{simonyan2014two}, both of which are computed from RGB frames. According to the way how dynamics are captured, the deep learning networks are roughly grouped into three categories. 
    
    The first group encodes motion cues of RGB frames in optical flows, and 2D CNNs are applied to build stronger representations. The two-stream framework \cite{simonyan2014two} is a representative, consisting of one 2D CNN that learns static spatial features and the other 2D CNN that models temporal information in optical flows, which are separately trained and then averaged for prediction. TSN \cite{wang2016temporal} enhances this two-stream fashion by sparse sampling and temporal fusion to improve motion features. 
    
    The second group represents RGB frames by 2D CNNs and calculates dynamics in the ConvNet feature space. One way is to launch Recurrent Neural Networks. For instance, \cite{yue2015beyond} exploits an independent LSTM network. Another choice is temporal convolution. \cite{jiang2019stm} shifts the features in the temporal dimension to optimize information exchange among neighboring frames, and \cite{jiang2019stm} designs separate modules to span the spatio-temporal feature space for encoding dynamics. \cite{li2020tea} develops distinct blocks to capture both the short and long range temporal evolutions. 
    
    The third group takes RGB frames as input and 3D CNNs are launched to extract unified spatio-temporal features. I3D \cite{DBLP:conf/cvpr/CarreiraZ17} inflates pre-trained 2D convolutions to 3D ones. To decrease heavy computations in 3D CNNs, \cite{tran2018closer} decomposes 3D convolutions into 2D spatial convolutions and 1D temporal convolutions or adopts a mixup of 2D/3D CNNs. SlowFast \cite{feichtenhofer2019slowfast} involves two 3D CNN paths, with the slow one to extract spatial semantics and the fast one to capture fine-grained motions.
    
    Despite competitive results, raw video based methods are limited by storage consumption and processing speed.
    
    \subsection{Action Recognition in Compressed Videos}
    
    Regarding action recognition in compressed data, I-frames (sparsely sampled RGB frames) and P-frames (\emph{e.g.} motion vectors and residuals) are used for feature extraction instead of all decoded RGB frames and computed optical flows. The early work \cite{zhang2016real} attempts to replace optical flows with motion vectors for a higher efficiency. CoViAR \cite{wu2018compressed} makes an extension by exploiting all the modalities, including I-frames, motion vectors, and residuals to bypass video decoding. To improve the dynamics, \cite{shou2019dmc} adopts GAN to refine motion vectors and \cite{cao2019compressed} applies image denoising techniques. \cite{huang2019flow} feeds concatenated motion vectors and residuals into a CNN to mimic a flow based teacher, and a similar idea is given by \cite{battash2020mimic}. \cite{huo2019mobile} presents Temporal Trilinear Pooling to integrate multiple modalities on lightweight models for portable devices. \cite{li2020slow} follows SlowFast and proposes the Slow-I-Fast-P model, and pseudo optical flows are estimated with a specific loss.
    
    The methods above boost the performance; however, they struggle with modeling coarse and noisy dynamics and fusing distinct modalities, which are addressed in this study. 
    
    \section{The Proposed Approach}
    
    As displayed in Fig. \ref{fig:2}, our proposed framework firstly applies encapsulation to extract I-frame and P-frame clips from compressed videos, where P-frames are given in Motion Vectors and Residuals, denoted as the MVR modality. Subsequently, MEACI-Net predicts the categories of actions presented in videos. 
    Concretely, MEACI-Net follows the classic two-stream architecture with the RGB (I-frame) and motion (MVR) streams. The former simply adopts I3D-ResNet50, and the later also introduces the same backbone but extends it as a novel Compressed Motion Enhancement (CME) network to handle coarse and noisy motion cues by employing a Multi-Scale Block (MSB) embedded with a Denoising Module (DM). 
    
    Moreover, MEACI-Net presents the Attentive Cross-modal Interaction (ACI) mechanism to strengthen information exchange between the two streams. ACI consists of several Selective Motion Complement (SMC) units in the low-level layers followed by a Cross-Modality Augment (CMA) unit in the high-level layer. SMC enhances representation learning in the RGB modality by incorporating informative motion cues from the MVR modality. CMA further builds a cross-modal representation as an augmentation of the single-modal ones by aggregating multi-modal features via cross attention. The high-level features extracted from the two individual streams together with the one learned by CMA are fused at the score level for final prediction. In the rest part, we elaborate the technical details of the main components. 
    
    \subsection{Compressed Video Encapsulation}
    
    The encoded compressed video is usually packed as multiple Group of Pictures (GOPs), each of which has an I-frame and several consecutive P-frames. The I-frame represents the reference RGB frame in each GOP, and the P-frames generally consist of the motion vectors and its residuals, where motion vector performs a coarse approximation on the movement of the target by encoding the displacement in each 16 $\times$ 16 macroblock relative to the I-frame and the residual is the difference between the raw RGB frame and the one reconstructed by the motion vector after motion compensation. Similar to \cite{wu2018compressed}, we calculate accumulated residuals and motion vectors, which are iterated to I-frames in GOPs. 
    Since motion vectors contain main dynamics and residuals include rich boundaries, we concatenate them as the input to the CME network. 
        
    \subsection{Compressed Motion Enhancement Network}
    
    As shown in Fig. \ref{fig:1}, motion vectors are coarse and noisy, making it difficult to learn discriminative temporal features for recognizing actions, especially to those similar ones such as painting lipstick and brushing teeth. It is therefore of critical importance to introduce multi-scale stimuli to explore more useful patterns. In the meantime, the noise in motion cues needs to be suppressed. To that end, inspired by \cite{li2020tea}, we present the Compressed Motion Enhancement (CME) network based on I3D-ResNet50 as in Fig.~\ref{fig:2}, which replaces the basic bottleneck block by a dedicatedly designed Multi-Scale Block, embedded with a Denoising Module (DM). 
    
    \begin{figure}[t]
        \centerline{\includegraphics[width=\linewidth]{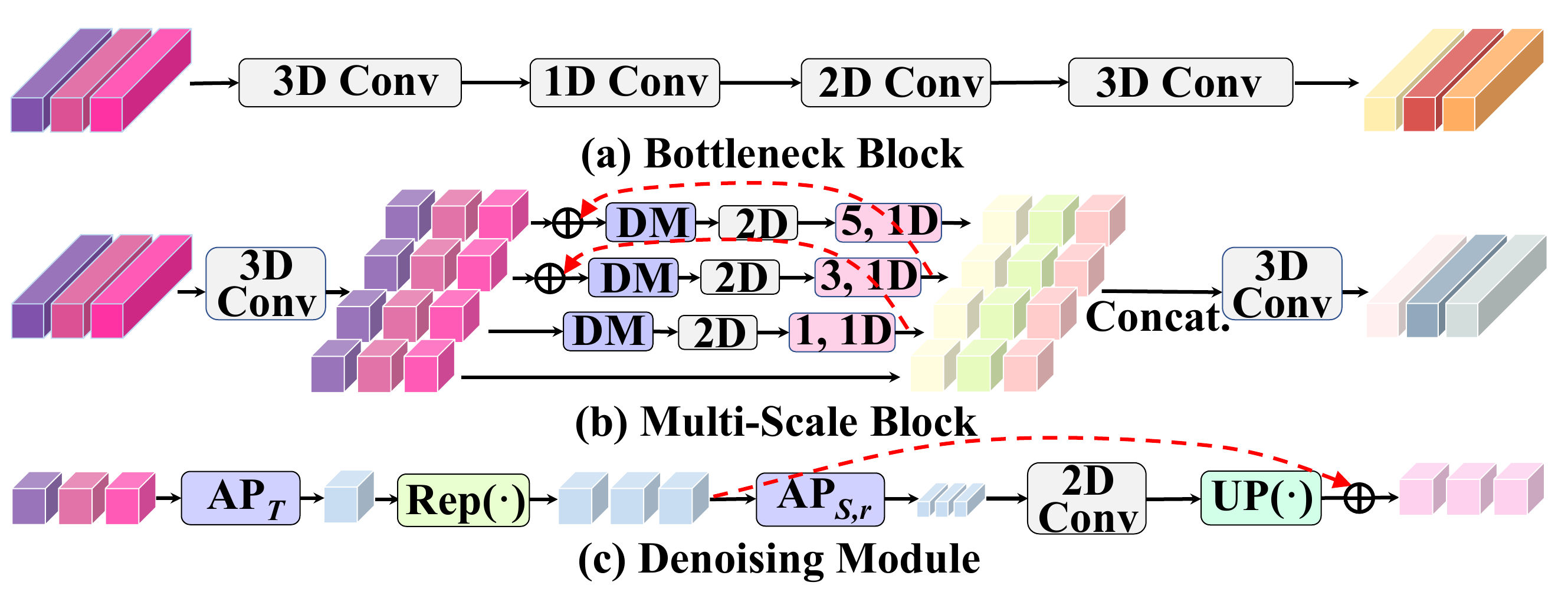}}
        \caption{Network structures: (a) the Bottleneck Block in I3D-ResNet50; (b) the Multi-Scale Block; (c) the Denoising Module.}\label{fig:3}
    \end{figure}
    
    \paragraph{The Multi-Scale Block (MSB).} Fig.~\ref{fig:3} (a) and (b) show the difference between the basic Bottleneck block used in I3D-ResNet50 and the Multi-Scale block proposed in this study. Instead of building features using a $3\times 1\times 1$ and a $1\times 3\times 3$ convolution kernel, MSB has four separate branches with cascaded connections, and short/long-term dynamics are captured by varying kernel sizes (\emph{i.e.} 1, 3, and 5), which efficiently extracts multi-scale motion patterns at multiple spatial granularities without adding too much extra computational cost. Specifically, MSB evenly splits the feature map after the first $1\times 1\times 1$ 3D convolution to four branches along the channel dimension, where the resulting group of feature maps is denoted by $\{\bm{X}_i\}_{i \in \{1,2, 3, 4\}}$. 
    Except the branch $\bm{X}_1$, for $i=2,3,4$, MSB removes the noise by the denoising module $DM(\cdot)$, leading to the refined feature map $DM(\bm{X}_{i})$, and performs a cascaded convolution $ST_{i}(\cdot)$ on $DM(\bm{X}_{i})$, where $ST_{i}(\cdot)$ is composed of a $3\times 3$ 2D spatial convolution followed by a 1D temporal convolution with kernel size ($2i-3$). Since $\{ST_{i}(\cdot)\}_{i=2,3,4}$ have different sizes of receptive fields, MSB encodes more abundant spatio-temporal patterns than the original Bottleneck block does. For the last two branches (\emph{i.e.} $i=3,4$), the output of the antecedent branch is also added into the input, \emph{i.e.} the input for $i=3,4$ is $\bm{X}_{i}+ST_{i-1}(DM(\bm{X}_{i-1}))$. This hierarchical design enables certain motion information to be captured at multiple scales. Note that it is not applicable to $i=2$, as $\bm{X}_{2}$ and the antecedent output $\bm{X}_{1}$ are at the same scale. Finally, to aggregate the multi-scale features, we concatenate all the output features from the four branches and fuse them by a $1\times 1\times 1$ 3D convolution. By this means, MSB represents coarse motion cues in a more comprehensive way, thus learning stronger features.  
    
    \paragraph{The Denoising Module (DM).} As in Fig.~\ref{fig:1}, the motion noise is always subtle and randomly appears at different positions and we suppress them in both the spatial and temporal domains. Inspired by \cite{moon2013fast}, it is promising to reduce this kind of motion noise by a multi-frame fusion. To that end, as shown in Fig.~\ref{fig:3} (c), $DM$ initially decreases the noise by aggregating multiple frames via temporal average pooling $AvgPool_{T}(\cdot)$, which is thereafter resized to the input size by a repeating operation $Rep(\cdot)$ formulated as below:
    \begin{equation}
        \bm{T_i} = Rep(AvgPool_{T}(\bm{X_i})).\label{eq:dm_t}
    \end{equation}
    Next, $DM$ spatially suppresses the noise by successively performing a spatial average pooling $AvgPool_{S,r}(\cdot)$, a $3\times 3$ 2D convolution $Conv_{3\times 3}(\cdot)$ and an up-sampling $UP(\cdot)$ based on bilinear interpolation as follows: 
    \begin{equation}
    \bm{S}_i = UP(Conv_{3\times 3}(AvgPool_{S,r}(\bm{T_i}))).\label{eq:dm_sp}
    \end{equation}
    Here, $AvgPool_{S,r}(\cdot)$ refers to average pooling with filter size $r\times r$ and stride $r=2^{i-1}$. In this way, the informative motion is saved while the subtle noise is simultaneously mitigated. 
    
    Finally, a skip connection as well as the Sigmoid function $\sigma$ are introduced to output the importance weights and the refined feature map is formulated as below:
    \begin{equation}
        DM(\bm{X}_{i}) = \sigma(\bm{T}_i + {\bm{S}_i})\odot \bm{X}_{i},\label{eq:dm}
    \end{equation}
    where $\odot$ denotes the element-wise product.
    
    
    \subsection{Attentive Cross-modal Interaction}
    In compressed videos, I-frames are sparsely sampled from raw RGB frames, with static appearances recorded. Despite being coarse and noisy, P-frames (\emph{i.e.} the motion vectors and residuals) contain complementary motion cues. It is therefore natural to consider the interaction between the features from I-frames and those from P-frames, to mutually enhance representation learning. However, most existing studies combine the features from the two modalities in a straightforward way, ignoring the issue of cross-modal misalignment caused by the distinct importance of spatio-temporal cues in I- and P-frames to action recognition. To address this, as in Fig.~\ref{fig:2}, we propose the Selective Motion Complement units at the low level and the Cross-Modality Augment unit at the high level, to perform cross-modal interaction learning for feature fusion. 
    
    \paragraph{The Selective Motion Complement (SMC) Unit.} Suppose $\bm{F}_{I,l}$ and $\bm{F}_{P,l}$ are the feature maps from the RGB  (I-frames) and MVR (P-frames) modalities in the $l$-th layer ($l=1,2,3,4$), respectively. The basic idea of SMC is incorporating aligned motion cues from the MVR modality into the RGB modality. To this end, we introduce a simple yet empirically effective unit by using attentions. Concretely, SMC firstly conducts spatio-temporal attention on $\bm{F}_{P,l}$
    \begin{equation}
        \bm{F}_{P,l}' = \bm{F}_{P,l} \odot \sigma(Att_{SP}(MP(\bm{F}_{P,l}))),
        \label{eq:smc_spt}
    \end{equation}
    where $MP(\cdot)$ is max pooling, and $Att_{SP}(\cdot)$ consists of two $1\times 1\times 1$ 3D convolution layers.
        
    \begin{table*}[!t]
        \centering
        \resizebox{\linewidth}{!}{
        \begin{tabular}{c|c|c|c|c|c|c|c|c}
        \hline
        \multicolumn{2}{c|}{Methods} & Reference & Input Size [MB] & GFLOPs & Optical flow & HMDB-51 & UCF-101 & Kinetics-400 \\
        \hline
        \multirow{10}{*}{\rotatebox{90}{\bf{Raw video based}}} & TSN & \cite{wang2016temporal} & 10.5 & 1600 & Train \& Test & 68.5 & 94.0 & 69.1 \\
            & I3D-RGB & \cite{DBLP:conf/cvpr/CarreiraZ17} & - & - & No & 74.8 & 95.6 & 71.1 \\
            & I3D+Flow & \cite{DBLP:conf/cvpr/CarreiraZ17} & - & - & Train \& Test & 80.7 & 98.0 & 63.4 \\
            & ARTNet & \cite{wang2018appearance} & 25 & 5875 & No & 70.9 & 94.3 & 70.7 \\
            & R(2+1)D+Flow & \cite{tran2018closer}  & 13.4 & 3040 & Train \& Test & 78.7 & 97.3 &72.0 \\
            & TSM & \cite{lin2019tsm} & 12 & 1950 & No & 73.2 & 96.0 & 74.1 \\
            & STM & \cite{jiang2019stm} & 12  & 2010 & No & 72.2 & 96.2 & 73.7 \\
            & SlowFast & \cite{feichtenhofer2019slowfast} & 30 & 1971 & No & 79.3 & 96.8 & 75.6 \\
            & TEA & \cite{li2020tea} & 12 & 2100 & No & 73.3 & 96.9 & 76.1 \\
            & TDN & \cite{wang2020tdn} & 18 & 3240 & No & 76.3 & 97.4 & 76.6 \\
        \hline\hline
        \multirow{13}{*}{\rotatebox{90}{\bf{Compressed video based}}} & EMV-CNN  & \cite{zhang2016real} & 6.5 & - & Train & 51.2 & 86.4 & -\\
        & DTMV-CNN & \cite{zhang2018real} & 6.5 & - & Train & 55.3 & 87.5 & - \\
        & CoViAR & \cite{wu2018compressed}  & 6.8 & 3615 & No & 59.1 & 90.4 & - \\
        & CoViAR + Flow & \cite{wu2018compressed} & 11.0 & 3970 & Train \& Test & 70.2 & 94.9 & -\\
        & CoViAR + PWC-Net & \cite{sun2018pwc} & 6.8 & - & Train & 62.2 & 90.6 & - \\
        & Refined-MV & \cite{cao2019compressed} & - & - & No & 59.7 & 89.9 & - \\
        & TTP & \cite{huo2019mobile} & 6.8 & 1050 & No & 58.2 & 87.2 & - \\
        & IP TSN & \cite{huang2019flow}  & 6.8 & 3400 & Train & 69.1 & 93.4 & - \\
        & DMC-Net (ResNet-18) & \cite{shou2019dmc} & - & - & Train & 62.8 & 90.9 & - \\
        & DMC-Net (I3D) & \cite{shou2019dmc} & - & 401 & Train & 71.8 & 92.3 & - \\
        & MFCD-Net & \cite{battash2020mimic} & 0.4 & 1300 & No &66.9 & 93.2 & 68.3 \\
        & SIFP-Net & \cite{li2020slow} & 8.1 & 1971 & No & 72.3 & 94.0 & - \\
        \cline{2-9}  
        & \textbf{MEACI-Net} (1-clip) &  Ours & 0.2 & 89 & No & \bf{74.0} & \bf{96.1} & 70.4 \\
        & \textbf{MEACI-Net} (3-clip) &  Ours & 0.7 & 268 & No & \bf{74.4} & \bf{96.4} & 71.5\\
        \hline 
        \end{tabular}
        }
        \caption{Comparison with the state-of-the-art approaches in the top-1 accuracy (\%) on the HMDB-51, UCF-101 and Kinetics-400 datasets. `MB' is short for MegaByte. `-' indicates that the corresponding result is NOT publicly available.}
        \label{tab:1}
    \end{table*}

    Afterwards, SMC applies channel-wise attention on the attended MVR feature $\bm{F}_{P,l}'$, which is further added to $\bm{F}_{I,l}$ from the RGB modality as follows:
    \begin{equation}
        \bm{F}_{I,l}: = \bm{F}_{I,l} + \bm{F}_{P,l}' \odot \sigma(Att_{C}(MP(\bm{F}_{P,l}'))),
        \label{eq:smc}
    \end{equation}
    where $Att_{C}(\cdot)$ is a $3\times 3\times 3$ 3D convolution layer.

    \paragraph{The Cross-Modality Augment (CMA) Unit.} To further enhance cross-modal interaction, CMA is presented to fuse the high-level features from the two modalities, which adopts a transformer-like structure. Specifically, CMA employs $\bm{F}_{I}$ and $\bm{F}_{P}$ after the last convolutional layer as the high-level features, based on which the key $\bm{K}_{m}$, query $\bm{Q}_{m}$ and value $\bm{V}_{m}$ are generated from $\bm{F}_{m}$ by linear mapping, and $m\in \{I,P\}$ indicates the modality. Thereafter, two features are learned for each modality by applying non-local cross attention as $\bm{F}_{I,att} = Softmax(\frac{\bm{Q}_{P}\bm{K}_{I}^T}{\sqrt{d_k}})\bm{V}_I$ and $\bm{F}_{P,att} = Softmax(\frac{\bm{Q}_{I}\bm{K}_{P}^T}{\sqrt{d_k}})\bm{V}_{P}$, where $d_{k}$ is the dimension of the key and $Softmax(\cdot)$ denotes the softmax function. CMA then sums the features as the combined representation across modalities as $\bm{F}_{fused}=\bm{F}_{I,att}+\bm{F}_{P,att}$. Finally, $\bm{F}_{fused}$ ensembles with the single-modal feature $\bm{F}_{I}$ and $\bm{F}_{P}$ at the score-level, yielding the overall prediction score for classification: $\bm{s} = \frac{1}{3}\cdot (CLS(\bm{F}_{I})+CLS(\bm{F}_{P})+CLS(\bm{F}_{fused}))$,
    where $CLS(\cdot)$ is the fully-connected classification head.
    
    For more details about the network structures of SMC and CMA, please denote the \emph{Supplementary Material}\footnotemark[2].

    \section{Experimental Results and Analysis}
    
    \subsection{Datasets and Evaluation Protocol}
    We adopt the following three benchmarks for evaluation. 
    
    \textbf{HMDB-51} contains 6,766 videos from 51 action categories and provides 3 training/testing splits. Each split consists of 3,570 training clips and 1,530 testing clips. \textbf{UCF-101} includes 13,320 videos from 101 categories. Similar to HMDB-51, this dataset also offers 3 training/testing splits, each of which has approximately 9,600 clips for training and 3,700 clips for testing. \textbf{Kinetics-400} is a large-scale dataset covering 400 categories, where 240k trimmed videos are used for training and 20k videos for validation.
    
    By following existing works, we report the top-1 accuracy on average over the three training/testing splits. 
            
    \subsection{Implementation Details}
    
    Similar to the case in the literature \cite{battash2020mimic}, we convert videos into the unified MPEG-4 Part2 coding format, where the size of GOPs is set as 12, and the frames are resized to 340$\times$256. By following \cite{li2020slow}, we pre-train the model on Kinetics-400 \cite{DBLP:conf/cvpr/CarreiraZ17}, and fine-tune it on HMDB-51 and UCF-101 by using the cross-entropy loss and the SGD optimizer with the weight decay of 0.0001, momentum of 0.9 and batch size of 36. The learning rate is initially set to 0.0001 and decreased by a factor of 10 for every 40 epochs. In training, we uniformly sample 8 frames to generate the input clip from each video and apply random scaling, corner crop and horizontal flip for data augmentation. All the experiments are conducted on two  NVIDIA 3090 GPUs. In testing, we uniformly sample 1 or 3 different clips from the input video and average the classification scores for final prediction when using 3 clips. Each frame is resized to 256$\times$256 and cropped into a 224$\times$224 image via center crop for inference.
    
    \subsection{Comparison with the State-of-the-arts}
    
    
    \paragraph{Comparison to Compressed Video based Methods.} As demonstrated in Table \ref{tab:1}, MEACI-Net outperforms the counterpart compressed video based methods by large margins. In particular, MEACI-Net promotes the second best method SIFP-Net by 1.7\% and 2.1\% in 1-clip test on HMDB-51 and UCF-101 respectively, thus reaching the new state-of-the-art. More importantly, MEACI-Net utilizes 21$\times$ less GFLOPs than SIFP-Net, highlighting its efficiency. By using more testing clips per video (\emph{i.e.} 3 clips), the accuracies are further boosted with more GFLOPs. When comparing CoViAR and CoViAR+Flow, we can observe that optical flows remarkably improve the performance, since they contain much more accurate motion information than coarse motion vectors. In this case, our method still outperforms CoViAR+Flow with 54$\times$ smaller input size and 44$\times$ less GFLOPs. Other methods such as EMV-CNN, DTMV-CNN, IP TSN, DMC-Net and MFCD-Net additionally adopt optical flows or raw videos to guide network training, but are clearly inferior to MEACI-Net. We also make evaluations on large-scale datasets (\emph{e.g.} Kinetics-400) to show its capacity. MEACI-Net achieves a get 2.1\% higher accuracy than MFCD-Net which is the one compressed video based method reporting performance on Kinetics-400. These results validate that our method explores motion information cues more comprehensively and performs cross-modal interaction more effectively, therefore reaching the state-of-the-art both in accuracy and efficiency.  
        
    \paragraph{Comparison to Raw Video based Methods.} As shown in Table~\ref{tab:1}, raw video based methods generally report higher accuracies than compressed video based counterparts, since raw videos convey much more complete motion information. However, MEACI-Net significantly reduces this gap in accuracy, and even outperforms some latest raw video based ones such as TEA. Unsurprisingly, MEACI-Net is much more efficient than raw video based methods, requiring at least 52$\times$ smaller input size and 16$\times$ less GFLOPs. 
    
    \subsection{Ablation Study}
    To evaluate the effectiveness of the proposed components of MEACI-Net, we extensively conduct the ablation study.
    
    \paragraph{Effect of Different Modalities.} We investigate the contribution of each modality to the performance of our method. When separately analyzing the RGB and MVR modality, we utilize the I3D-ResNet50 and CME network as the learning models, respectively. In the presence of both modalities, we apply the full model MEACI-Net. As summarized in Table \ref{tab:2},  when only using RGB or MVR, the accuracy is sharply decreased, since I-frames of the RGB modality are severely sparse, and P-frames of the MVR modality are coarse and noisy. Their combination via MEACI-Net clearly promotes the performance, showing that they contain complementary information for action recognition. 
    
    \begin{table}[!t]\tiny
        \centering
        \resizebox{\columnwidth}{!}{
        \begin{tabular}{c|c|c}
        \hline
        Modality & HMDB-51 & UCF-101 \\
        \hline
        RGB (+ I3D-ResNet50) & 66.5 & 91.8 \\
        MVR (+ CME) & 66.3 & 91.1 \\
        \hline
        RGB+MVR (+ MEACI-Net) & 74.0 & 96.1 \\
        \hline
        \end{tabular}}
        \caption{Comparison results (\%) by using different modalities.}
        \label{tab:2}
    \end{table}
    
    \begin{table}[!t]
        \centering
        \resizebox{\columnwidth}{!}{
        \begin{tabular}{c|c|c|c}
        \hline
        Methods & HMDB-51 & UCF-101 & Params. [M] \\
        \hline
        B1 & 59.5 & 87.5 & 27.5 \\
        B1 + MSB\textsuperscript{*} & 61.1 & 87.9 & 15.3 \\ 
        B1 + MSB & 61.6 & 88.6  & 15.3\\
        B1 + DM & 64.7 & 89.8 & 28.4 \\ 
        \hline
        CME Network & \bf{66.3} & \bf{91.1} & \bf{23.2} \\
        \hline
        \end{tabular}} 
        \caption{Ablation results (\%) on CME and its components MSB and DM, based on the I3D-ResNet50 baseline (`B1').}
        \label{tab:3}
    \end{table}
    
    \paragraph{Effect of CME (MSB and DM).} To validate the effectiveness of the proposed CME network and evaluate the impacts of the MSB and DM components, we adopt I3D-ResNet50 as the baseline, denoted by `B1' (`Baseline\_1'), and the other three compared baselines, \emph{i.e.} `B1+MSB', `B1+MSB\textsuperscript{*}' and `B1+DB', indicating the model which replaces the Bottleneck block by MSB with or without varying temporal kernel, and adding DM after the $1\times3\times3$ 2D convolution layer of the Bottleneck block, respectively. The CME network is the model by combining MSB and DM as shown in Fig.~\ref{fig:3}. As reported in Table \ref{tab:3}, MSB and DB clearly promote the accuracy of the I3D-ResNet50 backbone, since they separately handle the coarse and noisy motion vectors. Varying temporal kernel delivers a gain of 0.7\% on UCF-101 compared to that of a fixed size (\emph{i.e.} 3). A combination of them further boosts the performance. In fact, CME improves the accuracy of I3D-ResNet50 by 6.8\% and 3.6\% on HMDB-51 and UCF-101 respectively, while saving about 15\% in size compared to B1. We visualize the activation maps by Grad-CAM of B1, B1+DM and B1+DM+MSB (CME) in Fig. 4, and we can see that DM helps to focus more on motion areas (in red boxes) as it suppresses local noise by assigning bigger weights to foreground regions. For more visualization results, please refer to the \emph{Supplementary Material}\footnotemark[2]. 
    
    \begin{table}[!t]
        \centering
        \resizebox{0.85\columnwidth}{!}{
        \begin{tabular}{c|c|c}
        \hline
        Methods & HMDB-51 & UCF-101 \\
        \hline
        B2 &72.7 & 95.5 \\
        B2 + Add & 72.9 & 95.6 \\
        B2 + LA & 73.2 & 95.8 \\
        \hline
        B2 + SMC & 73.5 & 95.9 \\
        B2 + CMA & 73.7 & 96.0\\
        MEACI-Net (full model) & \bf{74.0} & \bf{96.1} \\
        \hline
        \end{tabular}}
        \caption{Ablation results (\%) on attentive cross-modal interaction and its components SMC/CMA. `LA' is short for lateral connection.}
        \label{tab:4}
    \end{table}
    
    \begin{table}[!t]
        \centering
        \resizebox{\linewidth}{!}{
        \begin{tabular}{c|c|c|c}
        \hline
        \multicolumn{1}{c|}{\multirow{2}*{Model}} & Raw video based & \multicolumn{2}{c}{Compressed video based}\\
        \cline{2-4}
        \multicolumn{1}{c|}{\multirow{1}*{}} & I3D & CoViAR & MEACI-Net \\
        \cline{3-4}
        \hline
        Preprocessing & 1093.9 & 13.4 & 7.7\\
        Model inference & 166.7 & 95.6 & 77.9 \\
        \cline{1-4}
        Full pipeline & 1260.6 & 109.0 & 85.6 \\
        \hline
        \end{tabular}}
        \caption{Comparison of inference time (in ms) per video.}
        \label{tab:5}
    \end{table}
    
    \begin{figure}[h]
        \centerline{\includegraphics[width=0.9\linewidth, height=1.7in]{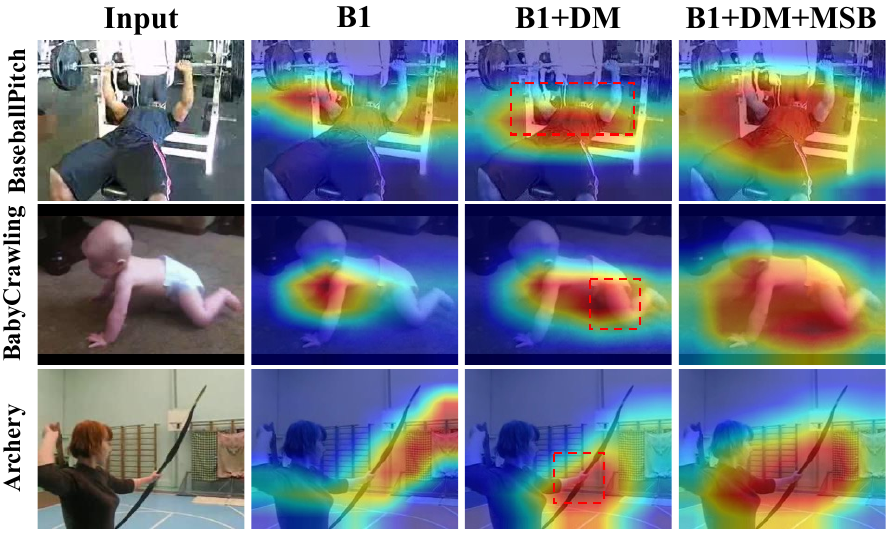}}
        \caption{Visualization of feature maps by Grad-CAM on UCF-101.}
        \label{fig:vis}
    \end{figure}
    
    \paragraph{Effect of ACI (SMC and CMA).} We compare the proposed ACI mechanism with other commonly used multi-modal fusion strategies. All the counterpart fusion techniques are implemented based on our two-stream framework as shown in Fig.~\ref{fig:2} with slight modifications. B2 (`Baseline\_2') is the baseline method without cross-modal interactions, \emph{i.e.} by removing the SMC and CMA units from MEACI-Net. `B2+Add' refers to the model directly adding the features of the two streams, acting like SMC but without attentions. Since SlowFast \cite{feichtenhofer2019slowfast} also considers cross-modal interaction by lateral connections, we therefore compare to it, denoted by `B2+LA'. `B2+SMC (CMA)' stands for the variant of our method by individually employing the SMC (CMA) unit. MEACI-Net is the full model by combining SMC and CMA. As shown in Table~\ref{tab:4}, all fusion techniques can improve the performance of the baseline, proving the necessity of cross-modal interactions. B2+SMC outperforms B2+Add, showing that low-level interaction across modalities facilitates multi-modal fusion. It is also shown that when high level interaction via CMA is integrated, the performance reaches the best. 
    
    \paragraph{Inference Speed.} As most methods do not report time costs or release codes, it is difficult to make fair comparison. In this case, we select open-sourced I3D and CoViAR as representatives of the raw and compressed video based methods. The experiments are conducted on a workstation with an Intel 2.3GHz CPU and an NVIDIA RTX 3090 GPU and the results are displayed in Table \ref*{tab:5}. With the proposed light-weight modules, MEACI-Net takes 85.6 ms for 8 sampled frames, \emph{i.e.} running at 93.4 FPS, faster than CoViAR and I3D.
    
    \section{Conclusion}

    In this paper, we propose an Attentive Cross-modal Interaction Network with Motion Enhancement (MEACI-Net) for compressed video action recognition. It employs the CME network to learn discriminative motion patterns from the coarse and noisy dynamics, and performs attentive cross-modal interaction for fusing features from multiple modalities at both the low and high levels. Extensive experimental results validate that MEACI-Net outperforms the state-of-the-art in both the accuracy and efficiency. 
    


\section*{Acknowledgments}
This work is partly supported by the National  Natural Science Foundation of China (No. 62022011), the Research Program of State Key Laboratory of Software Development Environment (SKLSDE-2021ZX-04), and the Fundamental Research Funds for the Central Universities.
\bibliographystyle{named}
\bibliography{references}

\clearpage

\appendix

\twocolumn[
\begin{@twocolumnfalse}
	\section*{\centering{Supplementary Material for \emph{Representation Learning for Compressed Video Action Recognition via Attentive Cross-modal Interaction with Motion Enhancement\\[10pt]}}}
\end{@twocolumnfalse}
]
\renewcommand\thefigure{\Alph{figure}}
\renewcommand\thetable{\Alph{table}}
\setcounter{table}{0}
\setcounter{figure}{0}

In this document, we provide more details about the proposed Selective Motion Complement (SMC) unit and the Cross-Modality Augment (CMA) unit in Section A, together with qualitative results of the proposed Compressed Motion Enhancement (CME) network in Section B.  

\section{Implementation Details of SMC and CMA}

\textbf{The SMC Unit.} Fig.~A (a) displays the detailed network structure of SMC. It first performs max-pooling $MP(\cdot)$ on the MVR feature $\bm{F}_{P,l}$, and then computes the spatio-temporal attention by the module $Att_{SP}$, which consists of two $1\times 1 \times 1$ 3D convolutional layers, and a RELU activation layer. Afterwards, SMC applies a channel-wise attention on the attended MVR feature $\bm{F}_{P,l}'=\bm{F}_{P,l} \odot \sigma(Att_{SP}(MP(\bm{F}_{P,l})))$, which is further incorporated to the RGB feature $\bm{F}_{I,l}$ via Eq.~(5) in the main paper. The specific architecture of SMC is summarized in Table A, which follows the format: $\#$kernel size, input channel size and output channel size.

\textbf{The CMA Unit.} CMA enhances cross-modal interaction by fusing high-level features $\bm{F}_{I}$ from the RGB modality and  $\bm{F}_{P}$ from the MVR modality. The non-local self-attention, which proves effective in exploring long-range dependencies, is employed, generating the weighted features $\bm{F}_{I,att} = Softmax(\frac{\bm{Q}_P\bm{K}_I^T}{\sqrt{d_k}})\bm{V}_I$ and $\bm{F}_{P,att} = Softmax(\frac{\bm{Q}_I\bm{K}_P^T}{\sqrt{d_k}})\bm{V}_P$ from the two modalities. Subsequently, CMA sums the features as the fused representation, which is further ensembled with the single-modal feature $\bm{F}_I$ and $\bm{F}_P$ at the score-level. The implementation details of CMA are summarized in Table B, following the same format as SMC: $\#$kernel size, input channel size and output channel size. The strides and sizes of padding of all the convolutional filters are set as 1.

\begin{figure}[!h]
    \centerline{\includegraphics[width=3.5in]{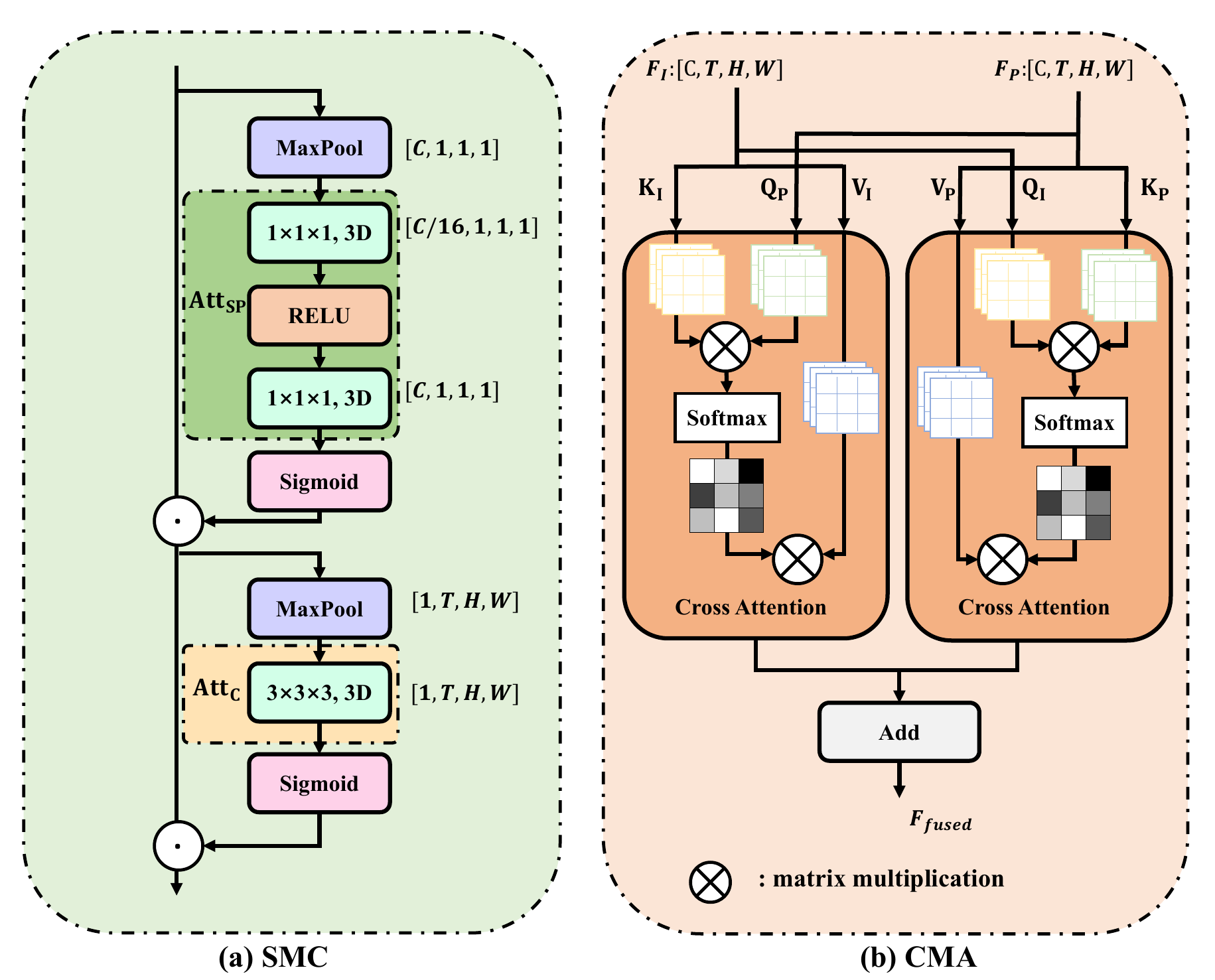}}
    \caption{Module details of (a) the Selective Motion Complement (SMC) unit and (b) the Cross-Modality Augment (CMA) unit.}
    \label{fig:4}
\end{figure}

\begin{table}[!t]
    \centering
    \resizebox{0.85\linewidth}{!}{
    \begin{tabular}{c|c|c}
    \hline
    \multicolumn{3}{c}{Network Configuration} \\
    \cline{1-3}
     Layer\_name & Layer~1 & Layer~2\\
    \hline
    conv0 & 1$\times$1$\times$1, 256, 16 & 1$\times$1$\times$1, 512, 32\\
    conv1 & 1$\times$1$\times$1, 16, 256 & 1$\times$1$\times$1, 32, 512\\
    conv2 & 3$\times$3$\times$3, 1, 1 & 3$\times$3$\times$3, 1, 1\\
    \hline
      Layer\_name & Layer~3 & Layer~4\\
    \hline
    conv0 & 1$\times$1$\times$1, 1024, 64 & 1$\times$1$\times$1, 2048, 128\\
    conv1 & 1$\times$1$\times$1, 64, 1024  & 1$\times$1$\times$1, 128, 2048\\
    conv2 &  3$\times$3$\times$3, 1, 1 & 3$\times$3$\times$3, 1, 1 \\
    \hline
    \end{tabular}
    }
    \caption{Architecture configuration of SMC.}
    \label{tab:arc_smc}
\end{table}

\begin{table}[!t]
   \centering
   \resizebox{\linewidth}{!}{
   \begin{tabular}{c|c}
   \hline
   Layer\_name & Network Configuration\\
   \hline
   $conv_{\bm{K}_I}$, $conv_{\bm{Q}_I}$, $conv_{\bm{V}_I}$ \\ $conv_{\bm{K}_P}$, $conv_{\bm{Q}_P}$, $conv_{\bm{V}_P}$  &
 $\left\{
       \begin{array}{l}
       {1\times1\times1, 2048, 128}\\
       {1\times1\times1, 2048, 128}\\
       {1\times1\times1, 2048, 128}\\
       \end{array} \right\} \times 2$
       \label{eq:msb} \\
   \hline
   \end{tabular}
   }
   \caption{Architecture configuration of CMA.}
   \label{tab:arc_cma}
\end{table}

\begin{figure}[!t]
    \centerline{\includegraphics[width=\linewidth]{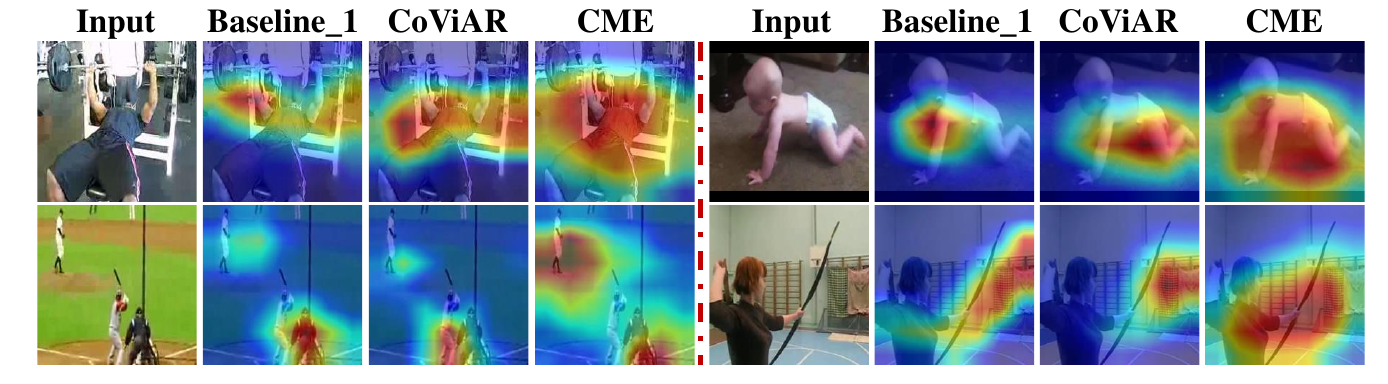}}
    \caption{Visualization of the activation maps by Grad-CAM of CME, I3D-ResNet50 (`Baseline$\_$1' or `B1' in the main paper) and CoViAR on UCF-101.}
    \label{fig:5}
\end{figure}

\section{Qualitative Results w.r.t CME}

We qualitatively demonstrate
the advantage of the proposed CME network by visualizing the class activation maps via Grad-CAM, compared to I3D-ResNet50 and CoViAR. As shown in Fig. B, CME clearly performs better in attending on discriminative regions for action recognition than I3D-ResNet50 and CoViAR. 

\end{document}